%% file: main.tex
\documentclass{article}
\usepackage{fullpage}

\input packages.tex
\input macros.tex

\usepackage{subfig}
\usepackage{xspace}
\usepackage{tikz}
\usetikzlibrary{arrows.meta,positioning,fit,calc,backgrounds}

\usepackage{pifont}

\newcommand{\OURS}{FlashNext-RL\xspace}
\newcommand{\qfn}{Qwen3.8-Flash-Next\xspace}
\newcommand{\primerl}{prime-rl\xspace}
\newcommand{\code}[1]{\texttt{#1}}

\definecolor{todoorange}{RGB}{230,120,0}

\newcolumntype{L}[1]{>{\raggedright\arraybackslash}p{#1}}
\usepackage{chngcntr}
\usepackage{adjustbox}

\newenvironment{titlemize}[1]{%
  \paragraph{#1}
  \begin{itemize}}
  {\end{itemize}}

\begin{document}

\title{
\OURS: A System Optimization Study for Long-Horizon Agentic RL of \qfn
}

\author{
Zhewei Yao, Yuxiong He
\\ Snowflake AI Research
}

\date{}
\maketitle

\input{_s0_abstract.tex}
\input{_s1_intro.tex}
\input{_s2_inference.tex}
\input{_s3_training.tex}
\input{_s4_e2e.tex}
\input{_s5_realrun.tex}
\input{_s6_conclusion.tex}

{
\bibliographystyle{plain}
\bibliography{ref.bib}
}

\end{document}

%% file: packages.tex
\usepackage[utf8]{inputenc} 
\usepackage[T1]{fontenc}    
\usepackage{url}            
\usepackage{booktabs}       
\usepackage{amsfonts}       
\usepackage{nicefrac}       
\usepackage{microtype}      
\usepackage{soul}
\usepackage{graphicx}
\usepackage{amsmath}
\usepackage{outlines}
\usepackage{xurl}

\usepackage[colorlinks=true,
citecolor=blue,
filecolor=black,
linkcolor=blue,
urlcolor=blue]{hyperref}

\input{math_commands.tex}

\usepackage{multirow}
\usepackage{paralist}

\usepackage{multicol}
\usepackage{diagbox}

\usepackage[ruled,noend]{algorithm2e}

\SetCommentSty{mycommfont}

\usepackage{here}

\usepackage{amsmath,amssymb,amsfonts,amsbsy,amsfonts,latexsym}
\usepackage{makecell}
\usepackage{xcolor}
\usepackage{colortbl}

\usepackage{tabularx,colortbl,xcolor}
\usepackage[normalem]{ulem}
\useunder{\uline}{\ul}{}

\usepackage{enumitem}

\usepackage{xparse}

\SetKwInput{KwInput}{Input}
\SetKwInput{KwRequire}{Require}

\usepackage{longtable}

%% file: math_commands.tex
\usepackage{amsmath,amsfonts,bm}

\usepackage{xcolor}
\usepackage{colortbl}

\def\eqref#1{equation~\ref{#1}}

\DeclareMathAlphabet{\mathsfit}{\encodingdefault}{\sfdefault}{m}{sl}
\SetMathAlphabet{\mathsfit}{bold}{\encodingdefault}{\sfdefault}{bx}{n}



%% file: macros.tex
\NewDocumentCommand{\var}{O{s} m O{}}{%
  \ensuremath{#1_{#2}^{#3}}
}
\usepackage{siunitx}

\newcommand{\commentout}[1]{}

\definecolor{light-gray}{gray}{0.80}

\newcommand\fref{Figure~\ref}
\newcommand\tref{Table~\ref}
\newcommand\sref{Section~\ref}

\usepackage{amsthm}

\usepackage{amsthm}

\newtheorem{mytheorem}{Theorem}
\newtheorem{assumption}[mytheorem]{Assumption}
\newtheorem{lemma}[mytheorem]{Lemma}
\newtheorem{remk}[mytheorem]{Remark}

%% file: _s0_abstract.tex
\begin{abstract}
Long-horizon agentic reinforcement learning (RL) requires both efficient rollout generation and training on long sequences.
However, the memory requirements and model-specific operators of \qfn make this difficult with existing open-source implementations.
To resolve these limitations, we introduce \OURS, a model-specific RL training optimization framework based on \primerl, with vLLM for software engineering (SWE) rollout generation.
Our main focus is BF16 training of the full 48-layer model on a single 8$\times$B200 node, with the first two layers, the n-gram table, the sparse-attention indexer and the multi-token prediction head frozen.
We optimize host offload, gradient accumulation and the kernels for mixture-of-experts, sparse attention and the gated residual stream.
In the full-model development study, non-padding training throughput increases from approximately 395 tokens/s/GPU in the initial 32K-token configuration to 4,438 tokens/s/GPU in the optimized 131K-token configuration, with hardware matmul throughput (aka without considering the cost of attention components) increasing from approximately 20 to 213 TFLOPs/GPU.
A controlled cumulative study on an 8-layer model further quantifies the contributions of individual optimizations.
Together with three inference nodes, the system supports a live SWE RL run with a target batch of 1,024 trainable rollouts and a 131,072-token context limit.
Over steps 5--15, an iteration takes 22.4 minutes on average, with bounded entropy and trainer--sampler mismatch KL.
These results highlight the importance of tailoring RL training to the model's architecture, memory layout and kernel requirements, rather than relying only on general-purpose RL infrastructure.
\end{abstract}

%% file: _s1_intro.tex
\section{Introduction}
\label{sec:intro}
Reinforcement learning (RL) is widely used in large language model (LLM) post-training to improve reasoning capabilities~\cite{shao2024deepseekmath,deepseekai2025r1,yu2025dapo}.
More recently, it has been applied to agentic tasks that require interaction with an environment over multiple turns~\cite{kimi2025k2,primeintellect2025intellect3}.
Software engineering (SWE) is a representative example: an agent reads a repository, executes commands and tests, and edits code to resolve an issue~\cite{jimenez2023swebench}.
SWE-Gym, R2E-Gym and SWE-smith provide environments or training data for developing such agents~\cite{pan2024swegym,jain2025r2egym,yang2025swesmith}, while SWE-rebench addresses task collection and decontaminated evaluation~\cite{badertdinov2025swerebench}.
In our setting, a rollout may contain many interaction turns, but its total context is capped at 131,072 tokens; a training batch contains approximately one thousand trainable rollouts.
Open-source RL systems, including DeepSpeed-Chat, OpenRLHF, HybridFlow and AReaL, address different aspects of training and generation~\cite{yao2023deepspeedchat,hu2024openrlhf,sheng2024hybridflow,fu2025areal}.
In this work, we build on \primerl for training and orchestration~\cite{primerl2025}, and verifiers for environments~\cite{verifiers2025}.

\qfn introduces architectural changes within the Qwen model family~\cite{yang2025qwen3,qwen2026design}.
The model configuration used in this study has the following components, consistent with the released configuration~\cite{qwen2026config}:
\begin{itemize}
    \item \textbf{Hybrid attention.} 36 of the 48 layers use Gated DeltaNet (GDN) linear attention~\cite{yang2024gated}. The other 12 use Qwen Sparse Attention (QSA), where a lightweight indexer selects up to 512 blocks of 4 keys per query. Related sparse-attention designs include NSA and DeepSeek Sparse Attention~\cite{yuan2025nsa,deepseekai2025v32}.
    \item \textbf{Gated residual.} A 4-branch residual stream ($4\times2560$) is mixed by learned read and write gates. Related multi-stream residual designs include hyper-connections and mHC~\cite{zhu2024hyperconnections,xie2025mhc}.
    \item \textbf{N-gram embedding.} A frozen 51B-parameter n-gram table supplies per-layer embeddings (PLE). Engram provides a related conditional-memory design~\cite{cheng2026engram}.
    \item \textbf{Fine-grained MoE.} Each layer has 512 routed experts with top-10 routing and an intermediate size of 640, plus a shared expert. This follows the broader use of sparse mixture-of-experts (MoE) layers~\cite{shazeer2017moe,deepseekai2024v3}.
    \item \textbf{Multi-token prediction (MTP).} An additional 4B-parameter head supports multi-token prediction, a technique studied in prior work~\cite{gloeckle2024mtp,deepseekai2024v3}. We use this head for speculative decoding~\cite{leviathan2023speculative,chen2023speculative}, rather than training it during RL.
\end{itemize}
In total, the model has 125B parameters (6B activated per token), plus the 51B n-gram table and the 4B MTP head.

However, the open-source implementations we evaluated did not meet our target of efficient training at a 131K-token context on one B200 node.
vLLM provides the serving backend~\cite{kwon2023vllm}, and training frameworks such as NeMo AutoModel and ms-swift provide useful infrastructure~\cite{nemoautomodel,zhao2024swift}.
Our target additionally requires model-specific backward kernels and memory planning across the RL loop.
In particular, we encountered three limitations:
\begin{enumerate}
    \item \textbf{GPU memory.} Without context parallelism, retaining one residual-stream input per layer for activation checkpointing~\cite{chen2016checkpointing} requires $48 \times 131{,}072 \times 10{,}240$ BF16 values, or about 129\,GB (120\,GiB) per GPU, before accounting for weights, gradients and layer-local intermediates.
    \item \textbf{Host memory.} With CPU offload~\cite{ren2021zerooffload,zhao2023fsdp}, storing an fp32 master parameter, an fp32 gradient and two fp32 AdamW moments requires 16 bytes per parameter. Our planning budget is about 1.9\,TiB before the n-gram table and the remaining RL system state, exceeding the node's host-memory capacity once these are included.
    \item \textbf{Kernels.} We found few efficient implementations of the backward pass of QSA's sparse grouped-query attention, of MoE kernels for an intermediate size of 640 at expert parallelism 8, or of fused kernels for the gated residual and the per-layer embeddings.
\end{enumerate}

To resolve those limitations, we introduce \OURS, which focuses on model-specific RL training optimization for \qfn in \primerl, with SWE as the example.
The RL setup uses four 8$\times$B200 nodes: one for training and three for inference.
We train the full 48-layer model, but freeze the first two layers, the n-gram table, the QSA indexer and the MTP head; this is not an update of every model parameter.
The contributions of \OURS are summarized as below.
\begin{enumerate}
    \item \textbf{Inference (\sref{sec:inference}).} A vLLM serving configuration for long agentic rollouts (per node, 2 engines of tensor parallelism 4 with expert parallelism; MTP speculative decoding; BF16 weights and KV cache).
    \item \textbf{Training (\sref{sec:training}).} A single-node BF16 trainer with CPU offload, streaming Muon~\cite{jordan2024muon,liu2025muon}, expert and context parallelism, and model-specific kernels. The optimized full-model development configuration reaches 4,438 non-padding tokens/s/GPU and 213 hardware matmul TFLOPs/GPU (\tref{tab:full_ladder}). Isolated comparisons and an 8-layer cumulative study quantify the contributions of individual optimizations.
    \item \textbf{End-to-end RL (\sref{sec:e2e}).} The memory, weight-transfer and checkpoint optimizations that let one training node run within a 2\,TB host-memory budget along with three inference nodes 
    \item \textbf{A real SWE RL run (\sref{sec:realrun}).} An end-to-end demonstration reporting iteration time, throughput, sequence lengths and short-run numerical behavior.
\end{enumerate}
Note that our goal is not to reach the best model accuracy or the most efficient RL recipe, but to identify the model-specific RL training optimizations that make long-horizon agentic RL of \qfn practical.

%% file: _s2_inference.tex
\section{Inference Optimization through vLLM Configuration Tuning}
\label{sec:inference}
We use vLLM for rollout generation~\cite{kwon2023vllm}; the version used in our experiments is 0.31 and supports \qfn.
This section describes serving configuration and the model-specific changes required by the RL loop, rather than a new inference backend.
Both the weights and the KV cache are in BF16.

\subsection{Topology}
\label{sec:inference_topology}
\paragraph{Expert parallelism.}
On B200, vLLM's BF16 MoE backend rounds the per-rank expert intermediate size up to a multiple of 128.
With tensor parallelism (TP) alone, each 640-wide expert is split across ranks: at TP 4, each rank holds 160 columns, padded to 256 (1.6$\times$), and the routed experts take 90.0\,GiB per rank.
With expert parallelism (EP), each expert stays whole and needs no padding: 56.25\,GiB per rank.
The padding reduces memory available for the KV cache and increases weight traffic for the selected experts during decoding.

\paragraph{TP degree.}
We replay real SWE rollouts turn by turn against each candidate engine, in a closed loop that reproduces the real prefix-cache behavior and MTP acceptance, and measure output tokens/s per GPU at a fixed number of concurrent sessions per GPU.
\tref{tab:inference_topology} shows the result.
TP 2 is the fastest at light load, but it has too little KV cache for SWE contexts of 64K tokens: beyond 10.5 sessions per GPU, its prefix cache stops hitting and it starts to preempt requests.
TP 8 has the most KV cache but the lowest throughput at every load.
We select TP 4 with EP for the target concurrency; it reaches 1,091 output tokens/s/GPU at 26 sessions per GPU in this replay.

\begin{table}[!htb]
\caption{
BF16 engine on B200 GPUs uses replayed SWE rollouts and MTP3. 
Output tokens/s per GPU; 
``KV'' is the KV-cache capacity per engine in tokens. 
The best measured configuration per row is in bold.
}\centering
\label{tab:inference_topology}
\begin{tabular}{lccc}
\toprule
Sessions/GPU & TP2 + EP (KV 1.76M) & TP4 + EP (KV 5.95M) & TP8 + EP (KV 8.41M) \\
\midrule
7    & \textbf{835} & 715 & -- \\
10.5 & \textbf{985} & 814 & -- \\
17.5 & 840 (prefix hit 0.06) & \textbf{943} & 658 \\
26   & 771 (318 preemptions) & \textbf{1,091} & 567 \\
35   & 783 (preempting) & \textbf{788} & 530 \\
\bottomrule
\end{tabular}
\end{table}

\subsection{Final Configuration}
Here is our final configurations used for the real training.
\label{sec:inference_config}
\begin{titlemize}{Engines.}
    \item \textbf{2 engines of TP 4 + EP per node}, i.e., 6 engines on the 3 inference nodes, on vLLM's V2 model runner. The V2 runner is required, because vLLM 0.31 captures routed experts (for router replay, below) with prefix caching on only in that runner.
    \item \textbf{MTP3 speculative decoding}, with the model's own MTP head as the draft model; it accepts 1.82--1.88 draft tokens per round on the replayed rollouts. The trainer does not update the MTP head, so its acceptance may decay over a long run, and we monitor it.
    \item \textbf{BF16 weights and KV cache}, with prefix caching on.
    \item \textbf{\code{gpu\_memory\_utilization} = 0.86}, about 5.6M KV tokens per engine. At 0.92, the KV cache is fully allocated at startup, and all 6 engines ran out of memory at the first weight update, when a 4.83\,GiB receive buffer met 3.3\,GiB of free memory (\sref{sec:weight_transfer}).
    \item \textbf{112 sessions per engine} (\code{max\_num\_seqs}), 672 rollouts in flight in total. In addition, the orchestrator stops admitting new sessions to an engine whose KV-cache usage passes 65\% (and resumes below 55\%), because the hybrid prefix cache stops hitting past about 70\% usage.
\end{titlemize}

\paragraph{One n-gram table per node.}
By default, each engine loads its own copy of the approximately 95.4\,GiB frozen n-gram table into pinned host memory, i.e., two copies per node.
We instead map a single copy per node from a node-local tmpfs and share it between both engines; the second engine then adds only 2.2\,GB.
The table must live on a tmpfs: \code{cudaHostRegister} refuses a writable shared mapping of a regular file, and the pod's \code{/dev/shm} is a 100\,GiB memory-backed volume that evicts the pod when it overflows.

\paragraph{Router replay.}
In BF16, two implementations of this model often pick different top-10 experts for the same token, and in our offline measurements these routing flips explain about 96\% of the trainer--inference mismatch.
The engines therefore return the routed experts of every token through vLLM's auxiliary-output connector ($[T, 48, 10]$ uint16, 960 bytes per token), and the trainer replays them~\cite{ma2025r3}.
The live-run mismatch KL is reported in \sref{sec:realrun}.

\subsection{Live Behavior}
\label{sec:inference_live}
In steps 5--15 of the live SWE run (\sref{sec:realrun}), engine telemetry reports 17.1K output tokens/s across the 6 engines: 2.8--2.9K per engine, 5.7K per node, or about 710 per GPU. 
\begin{itemize}
    \item \textbf{Sessions.} Of the 112 sessions per engine, about 77 are generating at any time; the others are between turns, waiting for a command to finish in their sandbox. Hardly any request waits for an engine slot (0.4--0.8 on average).
    \item \textbf{KV cache.} Usage averages 50\% and peaks at 72\%, so the admission cap engages only at the peaks; the prefix cache hits 85\% of prompt tokens.
    \item \textbf{MTP.} The mean acceptance length is 2.55, i.e., each forward pass of the target model emits 2.55 tokens on average.
\end{itemize}
At 672 rollouts in flight, about two thirds of the session slots are actively generating. Command execution in the sandboxes therefore contributes to unused serving capacity; these measurements do not isolate its contribution from GPU limits.

%% file: _s3_training.tex
\section{Training Optimization}
\label{sec:training}
Our main target is BF16 training of the full 48-layer model at a 131K-token bin on a \emph{single} 8$\times$B200 node.
We first describe the memory layout, then compare individual optimizations and evaluate their cumulative effect on an 8-layer model.
The full-model configuration freezes the components listed in \sref{sec:intro}; the 8-layer ladder does not freeze its first two layers.

\subsection{Setup}
\label{sec:train_setup}
\paragraph{Hardware.}
One node has 8 B200 GPUs (179\,GiB each), 192 CPU cores and about 2.0\,TB of host memory.
The four nodes are connected by 8$\times$400\,Gb/s EFA.

\paragraph{Data and metric.}
All throughput numbers in this section are measured by replaying recorded RL steps through the trainer, without generation and without the weight broadcast.
The batches are real SWE rollouts, packed into 131,072-token bins by the trainer's own packing.
Throughput measurements in the cumulative ladder disable PSS sampling, since walking process page tables measurably slows this offload-heavy workload. The separate memory ladder samples PSS every 30\,s; its step times are therefore indicative rather than directly comparable throughput measurements.
Generation is capped at the same length: the total context of a rollout (the prompt plus all turns) is at most 131,072 tokens, so a whole trajectory always fits in one training bin.
REAL tokens/s/GPU is the number of non-padding tokens in a step divided by the step time and by the 8 GPUs, averaged over the steady steps (step 1 includes compilation).
For model TFLOPs/GPU, we count forward and backward matmuls on each non-padding token's active path ($6P$ for a trainable matrix with $P$ weights), sparse-attention products, indexer scores and an estimate of the GDN recurrence, without activation-checkpoint recompute. On the replay batches, this is 40.3\,GFLOP per token for the unfrozen 48-layer model, 39.1 with the first two layers frozen, and 8.2 for the 8-layer model. Frozen matrices incur no weight-gradient cost, and output-head FLOPs are counted only at positions evaluated by the fused head. Hardware FLOPs additionally count checkpoint recompute, attention-backward score recompute and output-head logit recompute. Excluding attention, indexer scores and the GDN recurrence, the full model's hardware matmul cost is 48.4\,GFLOP per token with two layers frozen and 50.1 without.\footnote{The 8-layer model keeps the full 248K-vocabulary output head, which is 19\% of its FLOPs (4\% at 48 layers), so 8-layer and 48-layer TFLOPs/GPU are not comparable.}
The supplied FLOP accounting treats router weights and the QSA indexer as frozen. The output head uses fp32/TF32, and GDN retains an fp32 state; BF16 describes the main model compute rather than every operation. TFLOP rates are accounting estimates, not hardware-counter measurements.

\paragraph{The memory plan.}
\tref{tab:host_bytes} shows why a single node is hard.
Under CPU offload, the host holds fp32 master parameters, fp32 gradients and optimizer state. For a conservative planning budget, we use approximately 129B parameters before accounting for frozen components; \tref{tab:host_accounting} gives the measured final state.
With AdamW (two moments), this does not fit alongside the approximately 95.4\,GiB n-gram table and the remaining system state. Muon~\cite{jordan2024muon,liu2025muon} keeps one moment for its matrices and reduces the optimizer footprint.
The official model repository describes pretraining with Muon and AdamW~\cite{qwen2026design}, and we retain this division between parameter groups.
The checkpointed residual stream also leaves insufficient GPU memory for the remaining tensors under our unsharded 131K configuration (\sref{sec:intro}).
Our layout is:
\begin{itemize}
    \item \textbf{FSDP2 with CPU offload}~\cite{pytorchfsdp2}: parameters and optimizer state are offloaded to host memory, and FSDP units are gathered on the GPU as needed. Expert gradients accumulate on the GPU during a step before being offloaded (\sref{sec:train_gpugrads}). Prior work describes fully sharded training and CPU offload more broadly~\cite{zhao2023fsdp,ren2021zerooffload}.
    \item \textbf{StreamingMuon}: Muon's Newton--Schulz orthogonalization runs on the GPU, one parameter at a time, with each parameter streamed from host memory. Updates of an $A\times B$ matrix are scaled by $0.2\sqrt{\max(A,B)}$ so that their RMS matches AdamW's~\cite{liu2025muon}.
    \item \textbf{Expert parallelism 8}: each GPU owns 64 of each layer's 512 experts.
    \item \textbf{Ulysses context parallelism 4}~\cite{jacobs2023ulysses}: a 131K sequence is split over 4 GPUs (\sref{sec:train_cp}).
    \item \textbf{Full activation checkpointing}, with the forward pass's routing decisions reused in the recompute.
    \item \textbf{The first 2 layers frozen} (\sref{sec:train_freeze}).
\end{itemize}

\begin{table}[!htb]
\caption{Approximate host-memory budget for 129B parameters before layer freezing, on a node with about 1,937\,GiB available. The Muon row assumes one fp32 moment per parameter for budgeting; the implementation uses AdamW for the remaining parameter groups. Final measured state sizes are in \tref{tab:host_accounting}.}\centering
\label{tab:host_bytes}
\begin{tabular}{lccc}
\toprule
Optimizer & Bytes/param & Host for parameters & + n-gram table ($\sim$95.4\,GiB) \\
\midrule
AdamW & 16 & 1,922\,GiB & 2,018\,GiB (does not fit) \\
Muon budget & 12 & 1,442\,GiB & 1,537\,GiB (fits) \\
\bottomrule
\end{tabular}
\end{table}

\subsection{Isolated Optimizations}
\label{sec:train_isolated}
Each subsection below compares one technique with its alternative, with everything else fixed.
These comparisons were measured during development, on the code and data of that time; \sref{sec:train_ladder} measures all optimizations again in a single cumulative ladder on the same data.

\subsubsection{Host Offload: Pageable vs.\ PinnedArena}
\label{sec:train_arena}
FSDP2's pinned offload allocates from PyTorch's caching host allocator, which rounds every block up to a power of two (+28\% here) and never returns freed blocks; on the full model, the process was killed at 1,887\,GiB of host memory.
Pageable offload fits (1,555\,GiB peak) but copies at pageable speed: one StreamingMuon step takes about 75\,s.
Pinning the existing buffers in place with \code{cudaHostRegister} does not work either: once a registered region is freed and its address reused, the next registration fails and the rank dies.

\textbf{PinnedArena} maps one exact-size region per rank and registers it once; FSDP's flat parameter shards and StreamingMuon's moments are carved out of it, so no registered memory is ever freed.
A second exact-size \textbf{gradient arena} receives FSDP's gradient copies to the host (\sref{sec:host_mem}).
As \tref{tab:arena} shows, the Muon step drops from 66--75\,s to 12.4\,s.

\begin{table}[!htb]
\caption{Host offload path, full model on one node. Left: CP 1 with 32K bins (about 0.64M tokens per step). Right: CP 4 with a 131K bin.}\centering
\label{tab:arena}
\begin{tabular}{lccccc}
\toprule
& \multicolumn{2}{c}{CP 1, 32K} & \multicolumn{3}{c}{CP 4, 131K} \\
Offload & Muon step & Step & Muon step & Step & REAL tok/s/GPU \\
\midrule
Pageable & $\sim$75\,s & -- & 66\,s & 349\,s & $\sim$426 \\
PinnedArena & 20--28\,s & 161--179\,s & 28\,s & 288\,s & $\sim$516 \\
+ pinned gradients & 12.4\,s & 121--131\,s & 12.5\,s & 242\,s & 571--638 \\
\bottomrule
\end{tabular}
\end{table}

\subsubsection{Gradients: Blocking Host Copies vs.\ GPU-Resident Expert Gradients}
\label{sec:train_gpugrads}
With PinnedArena, a micro step of about 32K local tokens still took about 51\,s, against about 8\,s of compute.
A profile put 89.3\% of the autograd thread's samples inside FSDP2's \code{new\_sharded\_grad.to("cpu")}, with the GPUs at 7--8\% utilization.
From the second micro step on, whenever a gradient already exists on the host, FSDP2 copies the new one with \code{non\_blocking=False} and adds it on the host.
As a result, about 63\,GB of fp32 gradients per rank, mostly those of the EP-local experts, moved at 1.8\,GB/s in every micro step.

The EP-local experts are FSDP units of world size 1, so their sharded and unsharded gradients are the same tensor.
For these units, we skip the gradient sync on every micro step but the last: their gradients accumulate on the GPU and go to the host once per step, through the pinned non-blocking path.
The dense units (about 4\,GB per rank) keep an offloaded accumulator, whose copies we also make pinned and non-blocking, with the addition moved off the critical path.
\tref{tab:gpugrads} shows a 2.4--2.7$\times$ throughput improvement. The expert accumulator requires about 60\,GB per rank, although the measured peak increase is smaller because tensor lifetimes overlap differently.
On the final configuration, the dense accumulator gives another 4\% (3,233/3,277 $\rightarrow$ 3,383/3,372 REAL tokens/s/GPU).

\begin{table}[!htb]
\caption{GPU-resident expert gradients, full model, CP 4, 131K bin, 16-sample batches (about 1.1M tokens per step), steps 2 and 3.}\centering
\label{tab:gpugrads}
\begin{tabular}{lcccc}
\toprule
Gradient path & Micro step & Step & GPU peak (alloc.) & REAL tok/s/GPU \\
\midrule
FSDP2 host accumulation & $\sim$51\,s & 242 / 243\,s & 118.5\,GiB & 571 / 589 \\
GPU-resident expert gradients & 13--16\,s & 101 / 91\,s & 149.3\,GiB & 1,368 / 1,571 \\
\bottomrule
\end{tabular}
\end{table}

\paragraph{Overlapping the remaining host traffic.}
The offloaded expert parameters still travel from host to device at every layer.
Expert-weight prefetch copies these size-one shards on a separate stream, so that the copy overlaps the previous layer, and casts them to BF16 on a host thread first, which halves the bytes.
On the full model, this gives +3.0\% (4,067 $\rightarrow$ 4,190 REAL tokens/s/GPU), for 19\,GiB more host memory and a few GiB of GPU memory outside PyTorch's pool; on 8 layers, where the offload traffic is small, it is neutral ($-$0.6\%).

\subsubsection{Context Parallelism: Ulysses CP 4 vs.\ CP 8}
\label{sec:train_cp}
A 131K micro batch requires context parallelism (CP) in our memory layout. A ring-attention implementation for softmax attention does not directly implement the GDN recurrence, so we use a Ulysses-style layout~\cite{jacobs2023ulysses}: in each layer, an all-to-all switches from a sequence shard to a head shard and back (with the Q, K and V all-to-alls packed into one). The headwise GDN and PLE adaptations below are specific to our implementation.
\begin{itemize}
    \item \textbf{GDN} splits its 16 key and 48 value heads over the CP ranks; the short convolution and the per-head decay parameters are sliced per local head.
    \item \textbf{QSA attention} splits its 24 query heads; its 2 KV heads are replicated, since there are fewer of them than CP ranks.
    \item \textbf{The QSA indexer} has only 4 heads and sums its scores over them, so it is not split: each rank all-gathers the small indexer projections, computes the selection redundantly (22\,ms at 131K, \sref{sec:train_indexer}) and caches it for the recompute.
    \item \textbf{The PLE convolution} uses a sequence-to-channel all-to-all over its 10,240 channels, which avoids a halo exchange and the edge cases at packed-sequence boundaries.
    \item \textbf{MoE, shared expert, gated residual and norms} are token-local and run on the local shard.
\end{itemize}
CP 4 is faster than CP 8 (1,368/1,571 vs.\ 1,132/1,182 REAL tokens/s/GPU in the setting of \tref{tab:gpugrads}): every micro step has fixed costs, and CP 8 halves the number of tokens they are amortized over.
CP 8 peaks lower (124.9 vs.\ 149.3\,GiB allocated), but CP 4 fits.
Checkpointing several layers as one segment would also save memory, but it costs 11--22\% more backward time, so we do not use it.

\subsubsection{MoE: DeepEP vs.\ Mixture-of-Kittens}
\label{sec:train_mok}
The baseline MoE layer dispatches tokens with DeepEP~\cite{deepep2025} at EP 8 and runs the experts with grouped GEMMs.
Mixture-of-Kittens (MoK)~\cite{sul2026mok}, built on ThunderKittens~\cite{spector2024thunderkittens}, fuses dispatch, the expert GEMMs and combine into one persistent kernel, for both the forward and the backward pass, and overlaps communication with compute inside it.
As \tref{tab:mok} shows, it is 1.56--2.24$\times$ faster per layer.
It also peaks much lower (15.9 vs.\ 71.1\,GiB at 131K tokens per rank), although part of its symmetric-memory workspace may be allocated outside PyTorch's allocator and not counted.

Two further changes take another 22.7\% off the MoK layer:
\begin{itemize}
    \item \textbf{Native intermediate size 640.} Upstream MoK requires a multiple of 256 and therefore pads 640 to 768, i.e., 20\% more expert FLOPs. Our fork computes the tile counts over the intermediate dimension by ceiling division and lets TMA zero-fill the tail tile, so any multiple of 128 works.
    \item \textbf{More communication SMs.} At 32K tokens per rank, the layer is bound by dispatch and combine more than by the GEMMs (about 10 of 45\,ms), so giving the in-kernel all-to-all 52/52 SMs (forward/backward) instead of 24/28 matters more than the padding.
\end{itemize}
Relative to the padded baseline, the native size reduces layer time by 4.8\%; the subsequent SM change reduces the remaining time by 18.7\%, for a combined reduction of 22.7\%.
On the full model, the two together raise throughput by 5.3\% (3,405 $\rightarrow$ 3,586 REAL tokens/s/GPU).

\begin{table}[!htb]
\caption{MoE layer at EP 8. Left: forward + backward, DeepEP with grouped GEMMs vs.\ MoK (upstream build: intermediate size padded to 768, 24/28 communication SMs). Right: MoK at 32K tokens per rank, forward + recompute + backward.}\centering
\label{tab:mok}
\begin{tabular}{lccc}
\toprule
Tokens/rank & DeepEP & MoK & Speedup \\
\midrule
16K  & 32.2\,ms  & 20.7\,ms  & 1.56$\times$ \\
64K  & 130.1\,ms & 67.6\,ms  & 1.92$\times$ \\
131K & 294.0\,ms & 131.5\,ms & 2.24$\times$ \\
\bottomrule
\end{tabular}
\quad
\begin{tabular}{lcc}
\toprule
MoK variant & Comm SMs & Time \\
\midrule
I padded to 768 & 24/28 & 47.7\,ms \\
I native 640    & 24/28 & 45.4\,ms \\
I native 640    & 52/52 & 36.9\,ms ($-$22.7\%) \\
\bottomrule
\end{tabular}
\end{table}

\subsubsection{Static Expert Placement}
\label{sec:train_placement}
Expert load is skewed, and at EP 8 the busiest rank sets the pace of every MoE layer.
We record the expert loads of a few real steps and fit a static permutation that assigns experts to GPUs with the longest-processing-time-first heuristic.
Routes stay in logical expert ids; the MoE layer maps them to physical slots, and the weight broadcast and the HF export map the weights back to the logical order.
On held-out micro steps, the max/mean rank load drops from 1.233 (contiguous placement) to 1.059 on the full model, and from 1.131 to 1.030 on 8 layers.
The hottest expert carries at most 0.2 of a rank's mean load, so no expert needs replicas.
During development, placement gave +3.4\% on 8 layers (17,860 $\rightarrow$ 18,475 REAL tokens/s/GPU); on the final stack its effect is within 1\% (\tref{tab:train_ladder}).

\subsubsection{QSA Indexer: PyTorch vs.\ cuDNN Fused Indexer + Top-k}
\label{sec:train_indexer}
QSA's frozen indexer scores every 4-key block for each query and keeps the top 512.
cuDNN's DSA indexer kernels~\cite{cudnnfrontend} need 32 or 64 query heads per KV head, while QSA's indexer has 4, so we zero-pad the 4 heads to 32.
The fused cuDNN indexer + top-k is then 5.5$\times$ faster than the PyTorch implementation at 131K (\tref{tab:indexer}).
Because the indexer is frozen, the selection is computed once in the forward pass and reused in the recompute.

The cuDNN kernels compile one specialization per sequence length, which costs about 1\,s of host JIT time per new length while the other ranks wait; at 48 layers, this would have been about 24\,s per micro step.
We therefore pad each sequence to a power-of-two bucket (at least 4,096) by wrapping its own rows.
The padded blocks lie outside the causal window of every real query, so the selections are unchanged, and only about 6 buckets are ever compiled.
On 8 layers, this raised throughput from 8,243 to 13,571 REAL tokens/s/GPU during development.

\begin{table}[!htb]
\caption{QSA indexer + top-k, one B200, real layer inputs.}\centering
\label{tab:indexer}
\begin{tabular}{lcc}
\toprule
Sequence length & PyTorch & cuDNN fused \\
\midrule
8K   & 1.7\,ms   & 0.2\,ms \\
64K  & 35.3\,ms  & 6.1\,ms \\
131K & 123.8\,ms & 22.5\,ms \\
\bottomrule
\end{tabular}
\end{table}

\subsubsection{QSA Attention: Dense vs.\ cuDNN DSA vs.\ Triton vs.\ CuTe}
\label{sec:train_qsa}
\paragraph{Sparse attention is required.}
The original model code ran QSA as dense attention with FlashAttention~\cite{dao2022flashattention,dao2023flashattention2,shah2024flashattention3}, which computes a different function once a sequence is longer than 2,048 tokens, where the block selection starts.
Block-sparse masks do not help: on real data, a tile of 128 consecutive queries touches 90--94\% of the causal key tiles.
The trainer must therefore run token-level sparse attention, as inference does.

\paragraph{cuDNN DSA.}
The cuDNN sparse-attention kernels we evaluated~\cite{cudnnfrontend} target the MLA layout used in DeepSeek models~\cite{deepseekai2024v3,deepseekai2025v32}. Packing $[K|V]$ and $[Q|0]$ and zero-padding the 12 query heads of each KV group lets them compute our grouped-query attention, but the forward pass then leaves 52 of 64 head slots unused and the backward pass runs at only about 95\,TFLOP/s in our benchmark.

\paragraph{GQA-native Triton.}
Our Triton~\cite{tillet2019triton} kernel processes queries in windows: the 64 MMA rows hold the 12 heads of 5 consecutive queries, over the union of their selected blocks, with a per-query membership mask.
Neighboring queries select mostly the same blocks (the union for 5 queries is about 2$\times$ the blocks of one query, not 5$\times$), so gather traffic drops to about 0.41$\times$.
The dK/dV backward iterates over keys, using a key-tile CSR and a fixed-order reduction.
The kernel is 1.58$\times$ faster than DSA at 16K and 1.34$\times$ at 64K (\tref{tab:qsa}).

\paragraph{dK/dV.}
dK/dV then took 97 of about 130\,ms of the backward pass at 64K, bound by bandwidth from re-reading $q$ and $dO$ for every key tile.
We first moved the Triton gathers to TMA in a 3-stage pipeline (97 $\rightarrow$ 76.5\,ms) and then wrote a new CuTe-DSL~\cite{cutlass} kernel for sm100:
\begin{itemize}
    \item 64-key CSR tiles; each step processes 5 queries $\times$ 12 heads, i.e., 60 of the 64 MMA rows, gathered with \code{cp.async} directly into swizzled MMA tiles through a 2-stage ring;
    \item warp specialization: 4 gather warps (which also stage the per-row metadata), 4 compute warps for $P$ and $dS$, and 1 MMA warp;
    \item all accumulators in tensor memory: double-buffered $S$/$dP$ plus the $dV^T$/$dK^T$ halves, 512 columns in total;
    \item the MN-major operands are views of the gathered K-major tiles, so each row is gathered only once;
    \item the MMA warp issues the dV/dK update of step $i-1$ before $S$/$dP$ of step $i$, which frees a gather stage one phase earlier.
\end{itemize}
The kernel takes 46.7\,ms at 64K (1.63$\times$ faster than the Triton TMA version), close to both its gather floor and its MMA floor (about 41 and 40\,ms when each phase runs alone).
On the full model, the TMA gathers give +2.4\% and the CuTe kernel another +4.4\% (3,412 $\rightarrow$ 3,490 $\rightarrow$ 3,642 REAL tokens/s/GPU).

\begin{table}[!htb]
\caption{QSA attention, forward + backward per layer, real layer-7 selections, one sequence (one B200).}\centering
\label{tab:qsa}
\begin{tabular}{lccc}
\toprule
Implementation & 16K & 64K & dK/dV kernel at 64K \\
\midrule
cuDNN DSA (MLA kernels, padded heads) & 53.7\,ms & 219.8\,ms & -- \\
GQA-native Triton & 33.9\,ms & 164.5\,ms & 97\,ms \\
+ TMA gathers in dK/dV & 29.0\,ms & 138.5\,ms & 76.2\,ms \\
+ CuTe-DSL dK/dV (final) & 20.7\,ms & 106.4\,ms & 46.7\,ms \\
\bottomrule
\end{tabular}
\end{table}

\subsubsection{Gated Residual: Eager vs.\ Compiled vs.\ Fused}
\label{sec:train_hc}
Each layer reads the input of each of its two branches (attention and MLP) from the 10,240-wide residual stream, with a group norm, a down projection to 320 and a sigmoid-gated mix, and injects the branch output back: 96 read/inject pairs per forward pass.
Even compiled, these operations are memory-bound and keep large intermediates for the backward pass.
Our fused read and inject kernels (Triton plus cuBLAS) instead recompute the cheap intermediates in the backward pass.
\tref{tab:hc} shows a 1.95$\times$ speedup over \code{torch.compile} at 131K and 8$\times$ less memory kept per read/inject pair, which lowers the transient memory of each layer by about 10\,GiB.
In a real GDN decoder layer, forward + backward time drops by 31\%.

\begin{table}[!htb]
\caption{One gated-residual read + inject, forward + backward, one B200. ``Kept'' is the memory saved for the backward at 131K.}\centering
\label{tab:hc}
\begin{tabular}{lccc}
\toprule
& Eager & \code{torch.compile} & Fused \\
\midrule
16K  & 9.55\,ms  & 4.40\,ms  & 3.42\,ms \\
131K & 75.17\,ms & 34.74\,ms & 17.84\,ms \\
Kept at 131K & 15.79\,GiB & 5.79\,GiB & 0.71\,GiB \\
\bottomrule
\end{tabular}
\end{table}

\subsubsection{Other Fused Kernels}
\label{sec:train_glue}
\begin{titlemize}{}
    \item \textbf{PLE gate, norm and convolution.} A single packing-aware kernel, which restarts the convolution history at every packed-sequence boundary. It is 5.1$\times$ faster than the original eager path at 131K (70.71 $\rightarrow$ 13.92\,ms) and, unlike that path, does not convolve across sequence boundaries.
    \item \textbf{Attention q/k norm and partial RoPE.} A single Triton operator for the forward and backward pass: 2.28\,ms, against 13.5\,ms eager and 5.2\,ms compiled (32K tokens, 24 heads). It gives +6.4\% on 8 layers and +2.7\% on the full model.
    \item \textbf{Fused decoder glue.} Triton kernels for every RMSNorm, the QSA norm, RoPE and output gate, the PLE gate chain, and the decoder's shared-expert, routed and residual additions. On top of the q/k norm operator above, it gives +3.4\% on 8 layers (18,026 $\rightarrow$ 18,643) but only +0.7\% on the full model (4,406 $\rightarrow$ 4,438), whose step time is dominated by MoE, attention, GDN and offload.
\end{titlemize}

\subsubsection{N-gram Lookup: CPU Gather vs.\ UVA Gather}
\label{sec:train_ngram}
The 51B-parameter n-gram table (about 95.4\,GiB, or 102\,GB, in BF16) is frozen and lives in host memory.
The original code gathered its rows on the CPU and copied them to the GPU.
Under CP, the lookup runs on all 131K ids of the sequence, and the CPU \code{index\_select} left the GPU idle for about 540\,ms of every micro step (6.3\%).
Our UVA kernel lets the GPU read the rows directly from a host mapping of the table, on a side stream (\tref{tab:ngram}).
The mapping is a tmpfs file shared by all 8 ranks, since \code{cudaHostRegister} refuses a writable shared mapping of a regular file.
On the full model, this change alone gives +8.1\% (3,402 $\rightarrow$ 3,688 REAL tokens/s/GPU).

\begin{table}[!htb]
\caption{N-gram row gather for $T$ tokens ($[T, 16\times160]$ BF16 rows from the frozen table), one B200.}\centering
\label{tab:ngram}
\begin{tabular}{lccc}
\toprule
$T$ & CPU, 1 thread & CPU, 16 threads & UVA on GPU \\
\midrule
16K  & 40--42\,ms   & 14\,ms      & 5.2\,ms \\
131K & 395--464\,ms & 117--131\,ms & 42\,ms (no host cores) \\
\bottomrule
\end{tabular}
\end{table}

\subsubsection{Compilation and Stalls}
\label{sec:train_stalls}
With the 8 ranks synchronized by every MoE and FSDP collective, a stall on one rank stalls all of them.
\begin{titlemize}{}
    \item \textbf{Compilation under CP.} When CP is on, we compile~\cite{ansel2024pytorch2} only the token-local blocks (the gated residual, the shared expert and the final mixer); compiling whole layers split them into resume frames around the eager all-to-alls and broke the activation-checkpoint recompute.
    \item \textbf{Recompilation.} A \code{torch.compiler.disable} on the n-gram lookup caused a graph break whose resume frame recompiled for every new token count, so forward + backward time swung between 75 and 214\,s per step on 8 layers. Moving the lookup out of the compiled layer removed it.
    \item \textbf{Autotuning stragglers.} FLA's~\cite{fla2024} Triton kernels use the token count, rounded to 1,024, as an autotuning key. A rank whose micro batch fell into a new bucket re-tuned for seconds while the others waited; one rank was idle 45.5\% of the time. We remove the token count from those keys; together with other fixes made at the same time, 8-layer throughput went from 5,059 to 8,243 REAL tokens/s/GPU.
    \item \textbf{Pre-warming.} Compiling the dynamic-shape specializations and the indexer buckets at startup cuts step 1 from 171.9 to 23.8\,s; it does not change the steady state (\tref{tab:train_ladder}).
    \item \textbf{Copies and bookkeeping.} Removing a materialized transpose pair in GDN gives +6.8\% on 8 layers. Deferring the per-micro-step logging until after the last micro step raises the fraction of time the GPU is busy from 90\% to 96\% on the full model.
\end{titlemize}

\subsubsection{Freezing the First Two Layers}
\label{sec:train_freeze}
Freezing the 2 layers closest to the embedding removes their gradients, their optimizer state and their backward pass, including the largest single transient of the model: the backward pass of layer 1, the PLE layer, over the 10,240-wide stream.
On 8 layers with 64K tokens on one GPU, the activation peak drops from 107.6 to 78.6\,GiB.
On the full model, freezing is worth about 10\% (2,878/3,090 $\rightarrow$ 3,233/3,277 REAL tokens/s/GPU) and 37\,GiB of host memory (\sref{sec:host_mem}).

\subsection{Putting It Together}
\label{sec:train_ladder}
\begin{figure}[!htb]
    \centering
    \includegraphics[width=0.95\textwidth]{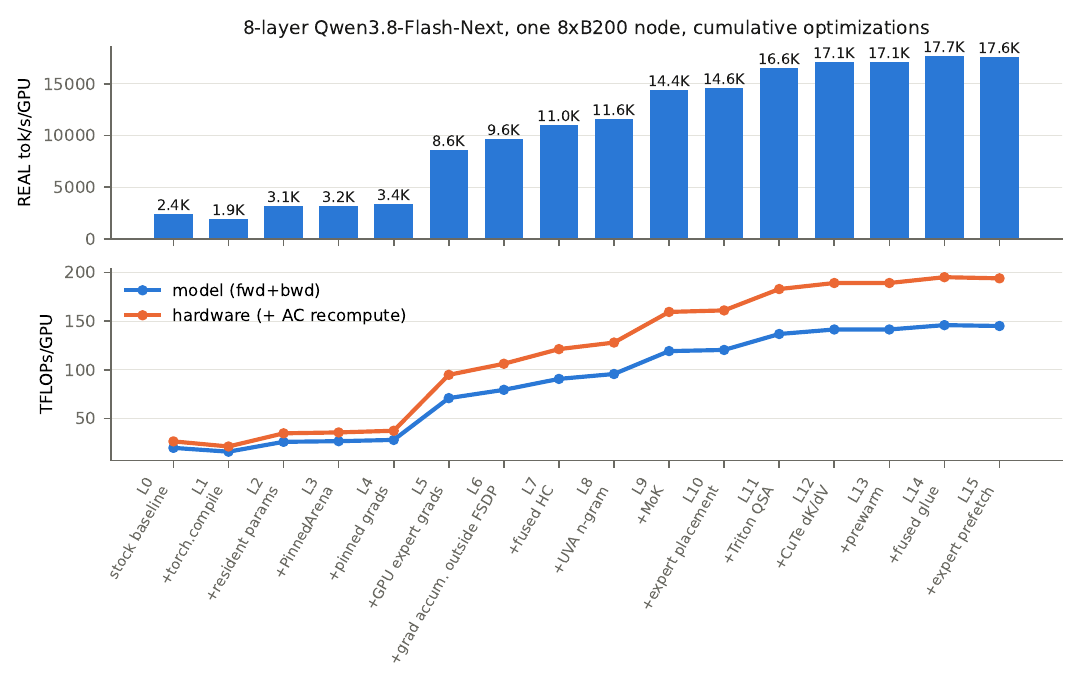}
    \caption{Cumulative trainer optimizations of \OURS on the 8-layer model: REAL tokens/s/GPU (top) and TFLOPs/GPU (bottom). Real rollouts, one node, 131K bins.}
    \label{fig:train_ladder}
\end{figure}

\paragraph{8-layer ladder.}
\fref{fig:train_ladder} and \tref{tab:train_ladder} add the optimizations one at a time to the first 8 layers of the model (6 GDN + 2 QSA layers, the same 3:1 mix as the full model, including the PLE layer and the n-gram table), on \qfn's own rollouts.
The baseline (L0) is the simplest configuration that runs at this shape: no compilation, DeepEP, QSA on cuDNN DSA, the eager gated residual, the CPU n-gram gather, pageable offload, and FSDP's default resharding after every forward pass.\footnote{A few fixes cannot be turned off and are in every rung: the indexer's power-of-two buckets, the packing-aware PLE convolution, the reuse of the forward routes in the recompute, the FLA autotuning fix, the removed GDN transpose, the packed Q/K/V all-to-all and the deferred logging.}
Throughput goes from 2,391 to 17,678 REAL tokens/s/GPU (7.4$\times$), and model TFLOPs/GPU from 20 to 146 (195 including the recompute); run-to-run noise is $\pm$0.3\%.
\begin{itemize}
    \item \textbf{The offload path comes first.} Keeping the parameters resident across micro steps (+64\%) and the expert gradients on the GPU (2.54$\times$) make up most of the gain: before these changes, the GPUs mostly wait for host copies.
    \item \textbf{Then come the kernels}: MoK (+24.6\%), the fused gated residual (+14\%), GQA-native QSA (+13.6\%), the UVA n-gram gather (+5.6\%), CuTe dK/dV (+3.4\%) and the fused glue (+3.2\%).
    \item \textbf{Order matters.} \code{torch.compile} costs 20\% on L0, where the step is bound by the eager gated residual and pageable copies, but gains 2\% on the final stack. Expert placement (+1\%), pre-warming (0\%, it only shortens step 1) and expert prefetch ($-$0.6\%) have small or negligible effects at 8 layers.
\end{itemize}
\tref{tab:remove_one} gives the complementary view, removing one optimization at a time from the final 8-layer stack:
MoK, the fused gated residual and the sparse attention kernels are each worth 15--22\%.

\begin{table}[!htb]
\caption{8-layer ladder (cumulative), mean of steps 2--7 on the same 8 batches. Model TFLOPs count the forward and backward passes; hardware (HW) TFLOPs add the recompute. Peak is rank 0's reserved GPU memory.}\centering
\label{tab:train_ladder}
\begin{adjustbox}{width=0.99\linewidth}
\begin{tabular}{llrrrrr}
\toprule
Rung & Change & REAL tok/s/GPU & vs.\ prev. & Model TFLOPs & HW TFLOPs & Peak GPU \\
\midrule
L0  & baseline (see text)                       & 2,391  & --         & 20  & 26  & 49.1\,GiB \\
L1  & + \code{torch.compile} (token-local blocks) & 1,916  & $-$19.9\% & 16  & 21  & 44.4\,GiB \\
L2  & + parameters resident across micro steps  & 3,142  & +64.0\%    & 26  & 35  & 52.5\,GiB \\
L3  & + PinnedArena                             & 3,225  & +2.6\%     & 27  & 36  & 52.8\,GiB \\
L4  & + pinned gradient copies                  & 3,381  & +4.8\%     & 28  & 37  & 52.7\,GiB \\
L5  & + GPU-resident expert gradients           & 8,587  & +154.0\%   & 71  & 95  & 64.9\,GiB \\
L6  & + dense gradient accumulation             & 9,618  & +12.0\%    & 79  & 106 & 65.0\,GiB \\
L7  & + fused gated residual                    & 10,981 & +14.2\%    & 91  & 121 & 61.4\,GiB \\
L8  & + UVA n-gram gather                       & 11,591 & +5.6\%     & 96  & 128 & 62.1\,GiB \\
L9  & + MoK (I = 640, 52/52 SMs)                & 14,445 & +24.6\%    & 119 & 160 & 49.1\,GiB \\
L10 & + static expert placement                 & 14,586 & +1.0\%     & 120 & 161 & 49.1\,GiB \\
L11 & + GQA-native Triton QSA                   & 16,573 & +13.6\%    & 137 & 183 & 49.2\,GiB \\
L12 & + CuTe dK/dV                              & 17,135 & +3.4\%     & 141 & 189 & 48.0\,GiB \\
L13 & + compile pre-warm                        & 17,138 & +0.0\%     & 141 & 189 & 48.9\,GiB \\
L14 & + fused decoder glue                       & 17,678 & +3.2\%     & 146 & 195 & 45.4\,GiB \\
L15 & + expert-weight prefetch                  & 17,570 & $-$0.6\%   & 145 & 194 & 47.3\,GiB \\
\bottomrule
\end{tabular}
\end{adjustbox}
\end{table}

\begin{table}[!htb]
\caption{Removing one optimization at a time from the final 8-layer stack (L14; 17,651 REAL tokens/s/GPU on the same node).}\centering
\label{tab:remove_one}
\begin{tabular}{lrr}
\toprule
Removed & REAL tok/s/GPU & Change \\
\midrule
MoK, back to DeepEP                      & 13,699 & $-$22.4\% \\
Fused gated residual and glue            & 14,191 & $-$19.6\% \\
GQA-native QSA, back to cuDNN DSA        & 15,028 & $-$14.9\% \\
UVA n-gram gather, back to CPU           & 16,058 & $-$9.0\% \\
MoK communication SMs, back to 24/28     & 16,845 & $-$4.6\% \\
CuTe dK/dV, back to Triton TMA           & 16,966 & $-$3.9\% \\
\code{torch.compile}                     & 17,319$^*$ & $-$2.0\% \\
Static expert placement                  & 17,632 & $-$0.1\% \\
\bottomrule
\multicolumn{3}{l}{\footnotesize $^*$Against L14 on its own node (17,678).}
\end{tabular}
\end{table}

\paragraph{Full model.}
\tref{tab:full_ladder} summarizes the full-model development measurements. The optimized 48-layer configuration reaches \textbf{4,438 REAL tokens/s/GPU} at a 131K-token bin, corresponding to \textbf{213 hardware matmul TFLOPs/GPU}. The initial configuration records approximately 395 REAL tokens/s/GPU at a 32K-token bin. These measurements document progress during development, rather than a controlled same-configuration speedup: the bin size, batch size and other settings change across rows.

\begin{table}[!htb]
\caption{Full 48-layer RL trainer development on one node. The initial row uses CP 1 and 32K bins; subsequent rows use CP 4 and 131K bins. Rows marked $^\dagger$ use 16-sample batches (about 1.1M tokens); later rows use the earlier rollout batches of 7.1--7.6M tokens (median sample 66K). HW TFLOPs count matmuls only (49.8\,GFLOP per token before freezing, 48.1 after, on these steps). Configurations and batch sizes vary, so the endpoints are not a controlled speedup comparison.}\centering
\label{tab:full_ladder}
\begin{tabular}{L{9.2cm}cc}
\toprule
Configuration & REAL tok/s/GPU & HW TFLOPs/GPU \\
\midrule
First full-model step (CP 1, 32K bins, pageable offload) & $\sim$395 & 20 \\
CP 4, 131K bin, pageable offload$^\dagger$ & $\sim$426 & 21 \\
+ PinnedArena$^\dagger$ & $\sim$516 & 26 \\
+ GPU-resident expert gradients$^\dagger$ & 1,368--1,571 & 68--78 \\
+ GQA-native Triton QSA (real steps from here on) & 2,297--2,333 & 114--116 \\
+ MoK, static placement, pre-warm, GDN transpose drop & 2,878--3,090 & 143--154 \\
+ freeze the first 2 layers & 3,233--3,277 & 156--158 \\
+ dense gradient accumulation & 3,412 & 164 \\
+ CuTe dK/dV, native MoK size + 52/52 SMs, UVA n-gram, q/k norm + RoPE, deferred logging & 4,406 & 212 \\
+ fused decoder glue & 4,438 & 213 \\
\bottomrule
\end{tabular}
\end{table}

%% file: _s4_e2e.tex
\section{End-to-End RL Optimization}
\label{sec:e2e}
Trainer and inference throughput alone do not determine RL performance.
The live loop must also buffer trajectories, transfer updated weights and save checkpoints within a limited memory budget.
This section describes the integration and memory optimizations needed to deploy the model-specific RL trainer.

\subsection{System Overview}
\label{sec:e2e_overview}
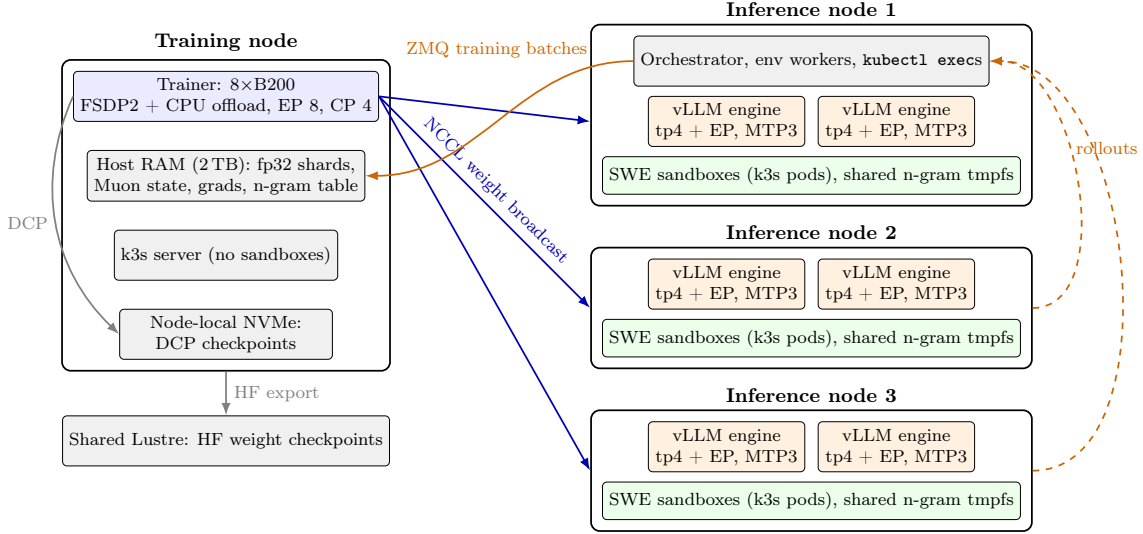
\begin{figure}[!htb]
    \centering
    \resizebox{0.95\textwidth}{!}{\input{figs/system.tex}}
    \caption{The RL setup of \OURS: one training node and three inference nodes, with the orchestrator on the first inference node. Blue: NCCL weight broadcast; orange: rollouts and training batches (with routed experts); gray: checkpoints.}
    \label{fig:system}
\end{figure}

\fref{fig:system} shows the layout of \OURS.
The RL setup uses four 8$\times$B200 nodes.
The training node runs the 8 trainer ranks and the control plane (k3s server) of the sandbox cluster, but no sandboxes.
Each of the three inference nodes runs 2 vLLM engines (\sref{sec:inference}) and its share of the SWE sandboxes, one Kubernetes pod per rollout.
The first inference node also runs the orchestrator and the environment workers (\sref{sec:host_mem_final}).
The orchestrator keeps 672 rollouts in flight, and the total context of each rollout is capped at 131,072 tokens, the trainer's bin size (\sref{sec:train_setup}).
A finished rollout returns its tokens and the routed experts of every token; the orchestrator assembles a batch and sends it over ZMQ to the training node, where rank 0 packs it into micro batches for all ranks.
After each step, trainer rank 0 broadcasts the new weights to all 24 inference GPUs with NCCL over EFA.
Generation and training overlap: the engines keep generating while the trainer works on the previous batch.

\subsection{Host Memory}
\label{sec:host_mem}
\paragraph{The budget.}
The training node has about 2.0\,TB of host memory, and the trainer alone needs most of it.
\tref{tab:host_accounting} gives the per-rank accounting of the final configuration: the 8 ranks hold about 1.4\,TiB of offloaded state, plus the 95\,GiB n-gram table that they share.
In our first configuration, the same node also ran the orchestrator, which buffers the rollouts with their routed experts (960 bytes per token, about 1\,GB per 32 rollouts), the environment workers, one \code{kubectl exec} process per running sandbox command (about 135\,MB each, up to about 90\,GB in total), and the k3s server.
We keep a host-memory safety margin on this node and stop the trainer when available memory falls below it, rather than let the kernel's OOM killer pick a victim.
Most of the fixes below started with a run or a replay crossing that margin.

\begin{table}[!htb]
\caption{Host memory per trainer rank (8 ranks on the training node), final configuration.}\centering
\label{tab:host_accounting}
\begin{tabular}{lcl}
\toprule
Structure & Size per rank & Where \\
\midrule
fp32 parameter shards (incl.\ the 2 frozen layers) & 58.9\,GiB & PinnedArena \\
StreamingMuon moments & 56.8\,GiB & PinnedArena \\
fp32 gradients & 56.1\,GiB & gradient arena \\
n-gram table (BF16) & 95.4\,GiB per node & shared tmpfs mapping \\
\bottomrule
\end{tabular}
\end{table}

\paragraph{Measure PSS, not RSS.}
Summing the per-process resident set size (RSS) overstates the trainer's memory by 43\%: 2,251\,GiB of RSS against 1,578\,GiB of proportional set size (PSS), because all 8 ranks map the shared n-gram tmpfs and each counts it in full.
Likewise, the pool of forked \code{torch.compile} workers shows 152\,GiB of RSS but only 5--8\,GiB of PSS (copy-on-write pages).
We therefore log PSS per process group every 10\,s and use it to attribute every drop in available memory.

\begin{figure}[!htb]
    \centering
    \includegraphics[width=0.98\textwidth]{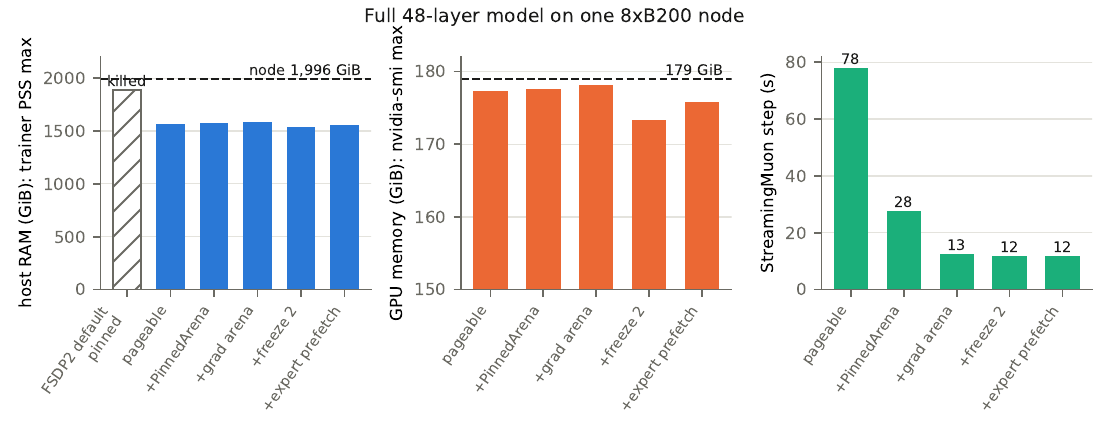}
    \caption{The full model on one node as the trainer's memory optimizations are added: trainer host memory (PSS), GPU memory and StreamingMuon step time. FSDP2's default pinned offload was killed before finishing a step.}
    \label{fig:mem_ladder}
\end{figure}

\paragraph{Memory optimizations.}
\tref{tab:mem_fixes} lists the changes and what each one saved; \fref{fig:mem_ladder} shows the host and GPU memory of the full model as the trainer-side optimizations are added.
\begin{titlemize}{Trainer state.}
    \item \textbf{Gradient arena.} FSDP2's pinned gradient copies came from the caching host allocator, rounded up to powers of two and never freed: 73.8\,GiB reserved per rank for 58.4\,GiB of gradients. The first gradient of each parameter in a step now lands in an exact-size buffer of a pinned arena, and later micro steps accumulate into it as before. Available memory rises by about 105\,GiB per node, with forward/backward and Muon times unchanged.
    \item \textbf{Re-freezing after EP.} Applying EP replaces the expert weights with new parameters, which default to \code{requires\_grad=True}, so the experts of the 2 frozen layers became trainable again. They never received a gradient, but they were given gradient buffers and optimizer moments. Re-freezing them after EP saves 37\,GiB per node.
    \item \textbf{Lazy optimizer state.} Allocating the Muon moments at construction put 450\,GiB of new pinned memory at trainer startup, exactly when the orchestrator starts its first 630 sandboxes (the environment workers grew from 2 to 236\,GB and \code{kubectl} from 0 to 80\,GB within 60\,s). The moments are now created at the first optimizer step.
\end{titlemize}
\begin{titlemize}{Batch path.}
    \item \textbf{Copies of the training batch.} A batch of 1,024 rollouts has about 39M tokens, and its routed experts alone take 960 bytes per token, about 37\,GB. The step boundary held two full batches: the previous step's micro batches stayed alive while the next batch was received and decoded. In addition, each rank widened the routed experts of its data-parallel group (about 19.5M tokens) to int32 on the host, about 37\,GB per rank or 300\,GB per node held for the whole step; each ZMQ frame was copied to \code{bytes} before decoding; and the orchestrator kept the routed experts of every buffered rollout twice, about 37\,GB more per batch of buffered rollouts. We now drop the previous batch first, keep the wire dtype on the host and widen on the GPU, decode in place, convert one micro batch at a time, and drop the per-turn copies of the routed experts once the training samples are built.
    \item \textbf{One CP shard per rank.} Every CP rank received the whole micro batch and kept the full routed-experts tensor on the host. Each rank now keeps only its own quarter of the rows: 720 bytes per token less, or about 14\,GB per rank and 110\,GB per node for a 39M-token batch.
    \item \textbf{Per-CP payloads.} Each rank still decoded the full payload of its data-parallel group. The packer now publishes one payload per (data-parallel rank, CP rank) pair. On a 16M-token test batch, the dip in available memory while the batch is received shrinks from 104--108\,GiB to 31--33\,GiB, and receiving and converting it takes 4.3--5.7\,s per rank instead of 9.3--9.8\,s; scaled linearly to a 39M-token batch, the dip shrinks from about 260\,GiB to about 80\,GiB.
\end{titlemize}
\begin{titlemize}{Sandbox setup.}
    \item \textbf{Node-local harness tarball.} Every sandbox setup read the 238\,MB agent-harness tarball into the environment worker's memory and piped it through its own \code{kubectl exec}; with hundreds of sandboxes starting at once, the environment workers reached about 200\,GB. The launcher now copies the tarball to the local disk of each node (checked by SHA-256), and each sandbox mounts it read-only through verifiers' \code{host\_files}~\cite{verifiers2025}. Sandbox setup drops from 17.9 to 4.7\,s (in a 20-sandbox test), the environment workers from 200 to 17.6\,GB, and the minimum available memory during a short test run rises from 82 to 289\,GiB.
\end{titlemize}
\fref{fig:mem_ladder} measures the trainer-side optimizations on the full model.
The exact-size arenas use no more host memory than pageable offload (trainer PSS of 1,567, 1,572 and 1,584\,GiB with pageable offload, PinnedArena and the gradient arena), whereas FSDP2's default pinned offload was killed at 1,887\,GiB.
What they buy is speed at the same footprint: the StreamingMuon step goes from 78 to 28 to 12.5\,s, and the full step from 415 to 330 to 216\,s.
Freezing the first 2 layers saves another 45\,GiB of host memory, and the expert prefetch costs 19\,GiB.
On the GPU, every configuration uses 173--178\,GiB of the 179\,GiB.
The checkpoint path had memory problems of its own, which we cover in \sref{sec:ckpt}.
With all of the above, the trainer's host memory stays flat across steps and checkpoints.

\paragraph{Batch 1,024: the final placement.}
\label{sec:host_mem_final}
With these fixes, available memory still crossed or approached the safety margin while a 39M-token batch was shipped, received or converted.
Three more fixes make each stage of the batch path hold the batch only about once:
\begin{itemize}
    \item \textbf{Streaming batch encode in the orchestrator.} When shipping a batch, the orchestrator held the training samples and a full encoding of the batch at the same time, +36\,GB within 15\,s. It now encodes the batch sample by sample into one buffer and frees each sample as it goes, with the same bytes on the wire; the peak for a full-size batch drops from 74.8 to 39.1\,GiB.
    \item \textbf{Rank-0 packer holds the batch once.} The packer peaked at about 2.6$\times$ the batch size (the received frame, the decoded batch, the micro-batch grid and all queued per-rank payloads), about 57\,GiB per receive. It now decodes the routed experts as views into the received frame, keeps them as uncopied chunks while packing, and keeps at most one payload in the send queue.
    \item \textbf{Per-micro-batch decode on every rank.} Each rank held its whole payload plus all decoded micro batches while converting them to tensors, about 75\,GiB across the 8 ranks for 1--2\,min per step. Micro batches are now decoded one at a time from zero-copy views and freed once converted, and the per-rank peak drops from 14.7 to 5.3\,GiB.
\end{itemize}
Finally, we moved the orchestrator, the environment workers, the model-request proxies and the \code{kubectl exec} processes to the first inference node, whose host memory is mostly unused.
In the live run, this stack uses about 190\,GB there (orchestrator 73--90\,GB, environment workers about 30\,GB, \code{kubectl} 76--96\,GB). Moving it was necessary for the target batch to fit within the training node's memory budget.
We also leave the expert prefetch of \sref{sec:train_gpugrads} off, because its 19\,GiB of host memory did not fit within the safety margin.
The live-run memory measurements are summarized in \tref{tab:realrun}; median available host memory is about 300\,GiB.

\begin{table}[!htb]
\caption{Memory optimizations on the training node.}\centering
\label{tab:mem_fixes}
\small
\begin{tabular}{L{4.2cm}L{6.2cm}L{4.8cm}}
\toprule
Optimization & Problem & Saved \\
\midrule
PinnedArena (\sref{sec:train_arena}) & Caching host allocator rounds up to $2^k$; in-place registration crashes & fits (default pinned offload killed at 1,887\,GiB) \\
Gradient arena & Pinned gradients rounded to $2^k$ and never freed & $\sim$105\,GiB per node \\
Re-freeze after EP & Frozen layers' experts got gradients and moments & 37\,GiB per node \\
Lazy optimizer state & 450\,GiB allocated during the sandbox startup burst & 450\,GiB off the startup peak \\
Batch copies & Two batches at the step boundary; int32 routes on the host; orchestrator copies & $\sim$300\,GB (int32 routes) + one batch + $\sim$37\,GB per buffered batch \\
CP host shard & Full routed experts on every CP rank & $\sim$110\,GB per node \\
Per-CP payloads & Every rank decodes the full payload & receive dip $\sim$260 $\rightarrow$ $\sim$80\,GiB \\
Node-local harness tarball & 238\,MB tarball piped per sandbox & environment workers 200 $\rightarrow$ 17.6\,GB \\
Streaming batch encode & Orchestrator holds samples + full encoding & $\sim$36\,GB at ship time \\
Rank-0 packer & Packer holds $\sim$2.6$\times$ the batch & $\sim$57\,GiB peak per receive \\
Per-micro-batch decode & Every rank holds payload + decoded batch & $\sim$75\,GiB across 8 ranks \\
Orchestrator on the first inference node & Orchestrator stack on the training node & $\sim$150--190\,GB \\
\midrule
Per-tensor arena storage (\sref{sec:ckpt}) & DCP cloned every arena view on save & $\sim$940\,GiB per save \\
N-gram table out of DCP & Every rank loaded the 95\,GiB table on resume & up to 760\,GiB on resume \\
Streamed HF export & Whole BF16 model gathered on rank 0 & $\sim$225\,GiB on rank 0 \\
\bottomrule
\end{tabular}
\end{table}

\subsection{Weight Transfer}
\label{sec:weight_transfer}
After every step, the trainer sends about 260\,GB of BF16 weights to the 6 engines.
The network is not the bottleneck: a node-to-node send over EFA runs at about 390\,GB/s, so the bytes alone take about 0.7\,s per receiving node.
The cost lies in gathering, converting and loading the weights, and in the memory each of these stages needs.
Our path is as follows:
\begin{itemize}
    \item \textbf{Sender, one layer at a time.} Trainer rank 0 builds an NCCL communicator with the 24 inference GPUs. For each layer, the trainer gathers the FSDP shards on the GPU, converts the parameter names to those of the HF checkpoint, undoes the static expert placement and restacks the 512 experts into the checkpoint's fused layout ($[512, 1280, 2560]$ and $[512, 2560, 640]$), so that each engine loads them with one call per shard.
    \item \textbf{No n-gram transfer.} The 95\,GiB n-gram table is never sent; each engine keeps its own copy, which vLLM's layerwise reload leaves untouched.
    \item \textbf{Bounded messages.} A layer is about 4.8\,GiB. We send it in buckets of at most 1\,GiB instead of as one message, which bounds the receiver's transient memory.
    \item \textbf{Receiver.} Each engine reloads layer by layer and yields clones rather than views of the receive buffer, so each buffer is freed as soon as its layer is loaded; with views, two partially loaded modules per layer kept every receive buffer alive until the engine ran out of memory.
\end{itemize}
Measured updates took 15--18\,s from the trainer's first send until the engines resumed. The live-run broadcast timings, reported separately in \tref{tab:realrun}, average 18.6\,s. Weight updates pause generation; training otherwise overlaps rollout collection.

\subsection{Checkpointing}
\label{sec:ckpt}
We save two kinds of checkpoints: a full DCP (PyTorch Distributed Checkpoint) checkpoint to resume from (fp32 parameters, all optimizer state, scheduler, progress and data loader; 923\,GB), and HF-format BF16 weights for serving and evaluation (about 225\,GB).
\paragraph{Full checkpoint.}
The DCP checkpoint goes to the training node's local NVMe rather than to the shared file system: a save takes about 70\,s, at the price of losing the checkpoint if the node is replaced.
Two fixes made it possible:
\begin{itemize}
    \item \textbf{Per-tensor arena storage.} DCP's writer clones any tensor whose storage is larger than the tensor itself and keeps the clones until the rank's file is written. Every arena tensor was a view into one arena-wide storage, so a save copied all parameter shards and moments, about 940\,GiB per node. Each arena tensor now wraps its own slice of the mapping (same address, still registered), and DCP writes directly from the arena.
    \item \textbf{The n-gram table out of DCP.} DCP saved the frozen table once and, on load, read the whole saved tensor on every rank, 95\,GiB per rank. The table never changes and lives in the shared mapping, so we leave it out of both save and load and only check on load that the mapping is populated.
\end{itemize}
Resuming from a live checkpoint reproduces the next live step.

\paragraph{Streamed HF weights.}
The original export gathered the whole BF16 model in rank 0's host memory before writing it, +225\,GiB.
Our streamed export all-gathers one layer at a time on the GPU, through the same gather-and-convert path as the weight broadcast, and rank 0 writes one file per layer from a writer thread.
Files that never change (the n-gram shards, the MTP head and the frozen layers) are linked from the base model, or written once per run and hardlinked into every step.
This keeps host memory flat, but the export still sits on the training path: with it, a checkpoint step cost 307\,s (71\,s for DCP and 236\,s for the streamed weights). Checkpoint cost depends on the model, not on the batch size.

\paragraph{HF weights from the broadcast.}
We therefore overlap part of the export with the weight broadcast and background disk writes. At a checkpoint step, rank 0 copies the layers that the weight broadcast has already gathered and converted into a small pool of pinned host buffers, and writer threads drain them to disk.
Only the layers that found no free buffer are gathered again, and the training step waits only while all buffers are busy.
In a trainer-only replay with 4 buffers, the export's time on the training path drops from 200--208\,s to 45--50\,s, with the same output as the streamed export.

In the real SWE RL run (\sref{sec:realrun}), the checkpoint at step 15 allocates 4 buffers of 4.89\,GiB once (5.1\,s); 20 of the 49 layer files come from the broadcast and 29 are gathered again.
The HF weights take 32.3\,s on the training path, the background write finishes 132.5\,s after it started, and the whole checkpoint step costs 102\,s instead of 307\,s (\tref{tab:ckpt}).
The checkpoint is complete: 1,658 tensors and 360\,GB of logical weights.

\begin{table}[!htb]
\caption{Checkpoint cost on the training path. $^*$The background write finished 132.5\,s after it started.}\centering
\label{tab:ckpt}
\begin{tabular}{lccc}
\toprule
Exporter & DCP & HF weights (on path) & Checkpoint-step overhead \\
\midrule
Streamed per layer & 71\,s & 236\,s & 307\,s \\
From the broadcast, 4 buffers & 70\,s & 32.3\,s$^*$ & 102\,s \\
\bottomrule
\end{tabular}
\end{table}

%% file: figs/system.tex
\begin{tikzpicture}[
  font=\footnotesize,
  box/.style={draw, rounded corners=2pt, align=center, inner sep=3pt, minimum height=0.8cm, minimum width=3.4cm},
  gpu/.style={box, fill=blue!8},
  cpu/.style={box, fill=gray!12},
  eng/.style={box, fill=orange!12, minimum width=2.5cm},
  sbx/.style={box, fill=green!8, minimum width=5.2cm, minimum height=0.6cm},
  grp/.style={draw, thick, rounded corners=4pt, inner sep=5pt},
  flow/.style={-{Latex[length=2mm]}, thick},
]
\node[gpu] (tr) at (0,0) {Trainer: 8$\times$B200\\FSDP2 + CPU offload, EP 8, CP 4};
\node[cpu, below=0.45cm of tr] (host) {Host RAM (2\,TB): fp32 shards,\\Muon state, grads, n-gram table};
\node[cpu, below=0.45cm of host] (k3s) {k3s server (no sandboxes)};
\node[cpu, below=0.45cm of k3s] (nvme) {Node-local NVMe:\\DCP checkpoints};
\node[grp, fit=(tr)(host)(k3s)(nvme), label={[font=\small\bfseries]above:Training node}] (tn) {};

\foreach \i/\y in {1/-0.4, 2/-3.0, 3/-5.6} {
  \node[eng] (ea\i) at (8.0,\y) {vLLM engine\\tp4 + EP, MTP3};
  \node[eng, right=0.2cm of ea\i] (eb\i) {vLLM engine\\tp4 + EP, MTP3};
  \node[sbx, anchor=north] (s\i) at ($(ea\i.south)!0.5!(eb\i.south) + (0,-0.15)$) {SWE sandboxes (k3s pods), shared n-gram tmpfs};
}
\node[cpu, minimum width=5.2cm, anchor=south] (orch) at ($(ea1.north)!0.5!(eb1.north) + (0,0.15)$) {Orchestrator, env workers, \texttt{kubectl exec}s};
\node[grp, fit=(orch)(ea1)(eb1)(s1), label={[font=\small\bfseries]above:Inference node 1}] (in1) {};
\node[grp, fit=(ea2)(eb2)(s2), label={[font=\small\bfseries]above:Inference node 2}] (in2) {};
\node[grp, fit=(ea3)(eb3)(s3), label={[font=\small\bfseries]above:Inference node 3}] (in3) {};

\node[cpu, below=0.7cm of tn.south, minimum width=4.4cm] (lustre) {Shared Lustre: HF weight checkpoints};

\draw[flow, blue!70!black] (tr.east) -- (in1.west |- ea1.west);
\draw[flow, blue!70!black] (tr.east) -- node[above, sloped, pos=0.5, text=blue!70!black] {NCCL weight broadcast} (in2.west);
\draw[flow, blue!70!black] (tr.east) -- (in3.west);
\draw[flow, orange!80!black] (orch.west) to[out=180, in=0] node[above, pos=0.12, anchor=south east] {ZMQ training batches} (host.east);
\draw[flow, orange!80!black, dashed] (in2.east) to[out=0, in=0, looseness=0.9] node[right, pos=0.6] {rollouts} (orch.east);
\draw[flow, orange!80!black, dashed] (in3.east) to[out=0, in=0, looseness=0.9] (orch.east);
\draw[flow, gray] (tr.west) to[bend right=35] node[left] {DCP} (nvme.west);
\draw[flow, gray] (tn.south) -- node[right] {HF export} (lustre.north);
\end{tikzpicture}

%% file: _s5_realrun.tex
\section{A Real SWE RL Run}
\label{sec:realrun}
We evaluate \OURS in a live SWE RL run, using steps 5--15 as the post-warmup reporting window.
The initial steps include compilation, optimizer-state allocation and sandbox startup, and are excluded from the performance summary.

\paragraph{Setup.}
The run uses the layout of \fref{fig:system}: one training node with the trainer configuration of \sref{sec:training}, except for the expert prefetch, whose 19\,GiB of host memory the node cannot spare (\sref{sec:host_mem_final}), and three inference nodes with the 6 engines of \sref{sec:inference} and 672 rollouts in flight.
The orchestrator runs on the first inference node.
The target batch is 1,024 trainable rollouts; collection can slightly exceed this target.
Groups of rollouts whose advantages are all zero (every rollout of a prompt gets the same reward) carry no learning signal and are not used for training. The orchestrator therefore keeps collecting until it has 1,024 trainable rollouts, so each step collects more: 1,249--1,584 rollouts in steps 5--15, 1,393 on average, of which 1,024--1,031 are trained (\tref{tab:realrun}).
A full checkpoint is written every 15 steps.

\paragraph{Speed.}
\tref{tab:realrun} gives the per-step breakdown.
One iteration takes 1,344\,s (22.4\,min) on average, and 1,137--1,690\,s across steps.
Of that time, the trainer works for 1,013\,s: 964\,s of forward and backward passes over about 36.5M packed tokens, 10\,s of data loading, 12\,s of StreamingMuon and 19\,s of weight broadcast.
Over the forward and backward passes, it processes 4,737 tokens/s/GPU (4,494--4,921), counting the packed 131K bins, i.e., about 229 hardware TFLOPs/GPU for the matmuls alone (\sref{sec:train_setup}); since this count includes the padding of the bins, it is not directly comparable with the REAL tokens/s/GPU of \sref{sec:train_ladder}.
For the remaining 331\,s (186--702\,s), the trainer waits for rollouts, so the loop is bound by generation.
The iteration-level generation rate averages 17.1K output tokens/s (\tref{tab:realrun}); engine-level telemetry is reported separately in \sref{sec:inference_live}. The checkpoint at step 15 adds 102\,s to that step.

\begin{table}[!htb]
\caption{The real SWE RL run, steps 5--15. ``Collected / trained'' counts rollouts; ``Wait'' is the trainer's wait for the batch; trainer tokens/s/GPU count the packed 131K bins over the forward and backward passes; ``Gen.'' is the number of output tokens generated per second of the iteration; ``Min avail.'' is the training node's minimum available host memory during the step. Step 15 includes a 102\,s checkpoint.}\centering
\label{tab:realrun}
\begin{adjustbox}{width=0.99\linewidth}
\begin{tabular}{rccccccccc}
\toprule
Step & Collected / trained & Iteration (s) & Wait (s) & Fwd/bwd (s) & Broadcast (s) & Trainer tok/s/GPU & Gen.\ tok/s & Min avail.\ (GiB) \\
\midrule
5 & 1,249 / 1,024 & 1,174 & 186 & 948 & 17.9 & 4,494 & 17.7K & 165 \\
6 & 1,356 / 1,028 & 1,260 & 276 & 943 & 17.2 & 4,700 & 17.9K & 161 \\
7 & 1,341 / 1,026 & 1,166 & 203 & 927 & 16.5 & 4,699 & 18.1K & 156 \\
8 & 1,264 / 1,024 & 1,137 & 196 & 904 & 16.6 & 4,812 & 17.7K & 159 \\
9 & 1,298 / 1,026 & 1,331 & 307 & 988 & 15.8 & 4,708 & 16.9K & 158 \\
10 & 1,351 / 1,031 & 1,306 & 328 & 940 & 18.0 & 4,739 & 16.9K & 153 \\
11 & 1,446 / 1,030 & 1,501 & 404 & 1,056 & 19.4 & 4,628 & 16.1K & 147 \\
12 & 1,378 / 1,024 & 1,288 & 233 & 1,014 & 20.0 & 4,719 & 18.1K & 140 \\
13 & 1,504 / 1,024 & 1,275 & 335 & 899 & 21.4 & 4,921 & 17.6K & 143 \\
14 & 1,547 / 1,024 & 1,690 & 702 & 951 & 17.2 & 4,893 & 14.6K & 153 \\
15 & 1,584 / 1,024 & 1,652 & 470 & 1,031 & 25.2 & 4,797 & 16.8K & 137 \\
\midrule
Mean & 1,393 / 1,026 & 1,344 & 331 & 964 & 18.6 & 4,737 & 17.1K & 152 \\
\bottomrule
\end{tabular}
\end{adjustbox}
\end{table}

\begin{figure}[!htb]
    \centering
    \includegraphics[width=0.95\textwidth]{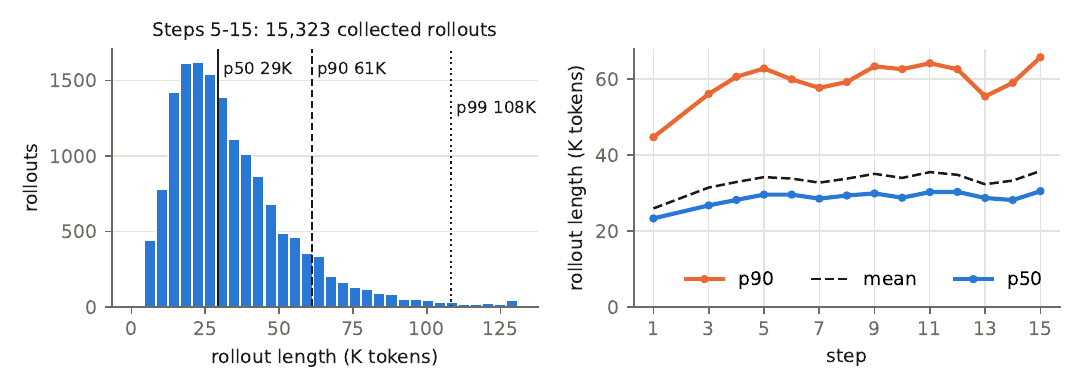}
    \caption{Rollout lengths in the real SWE RL run. Left: distribution over steps 5--15 (15,323 collected rollouts), with its median, p90 and p99. Right: median, mean and p90 per step (step 2's traces were not saved).}
    \label{fig:seqlen}
\end{figure}

\paragraph{Sequence lengths.}
\fref{fig:seqlen} shows the lengths of the collected rollouts.
In steps 5--15, the median rollout has 29K tokens, the p90 61K and the p99 108K, and some reach the 131,072-token cap; rollouts longer than 64K are 7.7\% of all rollouts but hold 19\% of the tokens.
Lengths grow over the first steps (the median goes from 23K at step 1 to about 29K at step 5) and then stay flat. 

\begin{figure}[!htb]
    \centering
    \includegraphics[width=0.95\textwidth]{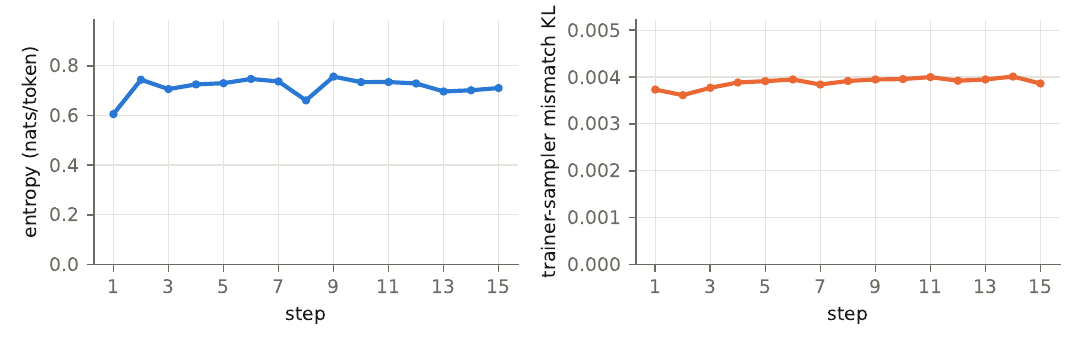}
    \caption{Entropy and trainer--sampler mismatch KL per step in the real SWE RL run.}
    \label{fig:entropy}
\end{figure}

\paragraph{Stability.}
\fref{fig:entropy} shows the entropy and the trainer--sampler mismatch KL per step.
After step 1, entropy remains within 0.66--0.76 nats per token, and mismatch KL with router replay remains within 3.6--4.0$\times10^{-3}$. These measurements show no numerical collapse over the observed run, but do not establish long-run convergence or task-quality improvement.
On the systems side, the minimum available host memory of the training node stays within 137--165\,GiB per step.

\paragraph{Illustrative capacity estimate.}
We estimate the additional training capacity needed to match the optimized trainer's service time, rather than claiming measured multi-node scaling or an exact end-to-end node requirement.
The estimate uses the following assumptions:
\begin{itemize}
    \item \textbf{Slowdown proxy.} The 8-layer ladder gives a 7.4$\times$ ratio (2,391 vs.\ 17,678 REAL tokens/s/GPU, \tref{tab:train_ladder}). Its baseline already includes mandatory correctness and compilation fixes, so it is not an unmodified stock implementation. The full-model development measurements at a 131K bin give a ratio of about 10.4$\times$ (426 vs.\ 4,438, \tref{tab:full_ladder}), but differ in batch size and configuration and are not a controlled speedup comparison.
    \item \textbf{Scaling.} Trainer throughput scales linearly with the number of training nodes. This is optimistic for the stock trainer, whose step was dominated by host copies.
    \item \textbf{Inference unchanged.} The generation side stays at three inference nodes.
\end{itemize}
Applying these slowdown proxies to all 1,013\,s of trainer work gives an illustrative requirement of 8--11 training nodes, or 11--14 nodes including inference, assuming ideal scaling.
The corresponding single-node service time would be 2.1--2.9\,h rather than 17\,min. This extrapolation treats fixed overheads as if they scaled with kernel throughput and is not a measured resource requirement; matching the 22.4-minute iteration time could require less training capacity because training overlaps generation.
Separately, the fp32 parameters, gradients and AdamW moments in the planning budget exceed one node's host-memory capacity when the n-gram table is included (\tref{tab:host_bytes}).

%% file: _s6_conclusion.tex
\section{Conclusion}
\label{sec:conclusions}
In this study, we explored model-specific RL training optimization for long-horizon agents using \qfn.
To address the model's memory and kernel requirements, we introduced \OURS, combining a single-node BF16 trainer with vLLM-based rollout generation.
Our experiments show that optimizing host transfers and gradient accumulation is necessary before kernel improvements can translate into end-to-end gains.
The live SWE run further demonstrates the importance of bounded batch buffering, layerwise weight updates and partially overlapped checkpoint export.
The resulting system supports a target batch of 1,024 trainable rollouts with one training node and three inference nodes.
The main lesson is that efficient RL training requires adapting the offload path, parallel layout and backward kernels to the model architecture; general-purpose RL infrastructure alone does not resolve these bottlenecks.
Our evaluation is limited to this hardware configuration and a short training run; it does not establish long-run convergence, final task accuracy or multi-node training scalability.
These remain important directions for future work.